\documentclass{article}

    \PassOptionsToPackage{numbers, compress}{natbib}
\usepackage{amsmath}

\usepackage[preprint]{neurips_2025}

\usepackage{tcolorbox}

\usepackage[utf8]{inputenc} % allow utf-8 input
\usepackage[T1]{fontenc}    % use 8-bit T1 fonts
\usepackage{hyperref}       % hyperlinks
\usepackage{url}            % simple URL typesetting
\usepackage{booktabs}       % professional-quality tables
\usepackage{amsfonts}       % Maroonboard math symbols
\usepackage{nicefrac}       % compact symbols for 1/2, etc.
\usepackage{microtype}      % microtypography
\usepackage[dvipsnames]{xcolor}         % colors
\usepackage{multirow} % For spanning rows
\usepackage{inconsolata}
\usepackage{enumitem}
\usepackage{pifont}
\usepackage{svg}
\tcbuselibrary{breakable}

\usepackage{float}
\newcommand{\CARE}{{\ttfamily CARE}}

\makeatletter
\renewcommand\paragraph{\@startsection{paragraph}{4}{\z@}%
  {.0ex \@plus 0.2ex \@minus 0.2ex}%
  {0pt}%
  {\normalfont\normalsize\bfseries}}
\makeatother

\let\oldparagraph\paragraph
\renewcommand{\paragraph}[1]{\oldparagraph{#1}~}

\title{CARE: Turning LLMs Into Causal Reasoning Expert}

\author{%
  Juncheng Dong\thanks{Equal contribution, first authors listed alphabetically.}\\
  Duke University\\
  Durham, NC \\
  \texttt{juncheng.dong@duke.edu} \\
  \And
  Yiling Liu\footnotemark[1] \\
   Duke University \\
   Durham, NC \\
  \texttt{yiling.liu@duke.edu} \\
  \And
  Ahmed Aloui \\
  Duke University\\
  Durham, NC \\
  \texttt{ahmed.aloui@duke.edu} \\
  \And
  Vahid Tarokh\thanks{Corresponding author.} \\
  Duke University\\
  Durham, NC \\
  \texttt{vahid.tarokh@duke.edu}
  \And David Carlson\footnotemark[2] \\
  Duke University\\
  Durham, NC \\
  \texttt{david.carlson@duke.edu} \\
}

\begin{document}

\maketitle

\begin{abstract}
Large language models (LLMs) have recently demonstrated impressive capabilities across a range of reasoning and generation tasks. However, research studies have shown that LLMs lack the ability to identify causal relationships, a fundamental cornerstone of human intelligence. We first conduct an exploratory investigation of LLMs’ behavior when asked to perform a causal-discovery task and find that they mostly rely on the semantic meaning of variable names, \emph{ignoring the observation data}. This is unsurprising, given that LLMs were never trained to process structural datasets. To first tackle this challenge, we prompt the LLMs with the outputs of established causal discovery algorithms designed for observational datasets. These algorithm outputs effectively serve as the sufficient statistics of the observation data. However, quite surprisingly, we find that prompting the LLMs with these sufficient statistics \emph{decreases} the LLMs' performance in causal discovery. To address this current limitation, we propose CARE, a framework that enhances LLMs’ causal-reasoning ability by teaching them to effectively utilize the outputs of established causal-discovery algorithms through supervised fine-tuning. Experimental results show that a finetuned Qwen2.5-1.5B model produced by CARE significantly outperforms both traditional causal-discovery algorithms and state-of-the-art LLMs with over a thousand times more parameters, demonstrating effective utilization of its own knowledge and the external algorithmic clues.

\end{abstract}

\section{Introduction}

Large Language Models (LLMs) have achieved remarkable success across diverse application domains, demonstrating astonishing capabilities in medicine~\citep{thirunavukarasu2023large}, natural language understanding \citep{achiam2023gpt}, complex code generation~\citep{chen2021evaluating}, and even accelerating scientific discovery~\citep{zhang2024comprehensive}. While these advancements reflect substantial progress in language understanding and reasoning, a critical examination reveals persistent challenges when LLMs perform causal discovery \citep{cai2023knowledge}.
Causal discovery, which infers causal relationships and structures from data, is fundamental to scientific understanding and human reasoning \citep{pearl2009causality, spirtes2000causation}. Despite this, there is an opportunity to use LLMs to improve traditional causal discovery algorithms that use data alone, as they can significantly benefit from external prior knowledge to reduce the hypothesis space, especially with limited data. Thus, the extensive world knowledge embedded in LLMs presents a rich resource for providing crucial guiding priors~\citep{darvariu2024large, hasan2022kcrl}.

However, although LLMs can sometimes generate plausible narratives about potential causal links, their capacity for accurately identifying causal graphs and distinguishing direct from indirect causes based on provided data frequently falls short~\citep{ashwani2024cause,long2023can}.
Specifically, recent studies show that LLMs often lack genuine causal reasoning capabilities \citep{chi2024unveiling,jin2023cladder,zevcevic2023causal}. Rather than conducting genuine causal thinking, they frequently engage in \emph{causal mimicry}: relying on recalling memorized facts, associative patterns, or plausible structures learned during pretraining on vast amounts of real-world causal information, and simply paraphrasing the knowledge contained in its pretraining corpora. This limitation becomes particularly apparent in data-driven tasks like causal discovery. 
% For example, when tasked with discovering a gene regulatory network (A, B, C) from provided expression data, a standard LLM might propose a biologically plausible structure reflecting known pathways from its training corpus. 
For example, when tasked with discovering the causal relationships between {\sc income}, {\sc education}, {\sc smoking}, and {\sc life expectancy} from a given observational dataset, LLMs mostly rely on the semantic meaning of these four variables and recite facts they have memorized during pretraining, such as ``{\sc smoking} causes {\sc life expectancy}'' and ``{\sc education} causes {\sc income}''. This behavior is the {\bf opposite} of conducting {\bf genuine} causal discovery based on the given observational data as done by established causal discovery algorithms such as \texttt{LiNGAM}~\citep{shimizu2006linear,shimizu2011directlingam}. 
% \emph{However, LLMs fundamentally struggles to effectively utilize the provided observational dataset.} 
In a recent work~\citep{cai2023knowledge} and Section~\ref{sec:investigation}, we showcase that \emph{\color{Maroon}LLMs fail to effectively utilize the given observation dataset}, mainly relying on the semantic meaning of variable names for causal discovery. Consequently, while performance might appear competent when variable names offer strong semantic clues, their discovered causal structures are often unreliable when genuine analysis of novel or complex data dependencies is required.

\begin{figure}[t]
  \centering
  \includegraphics[width=\linewidth]{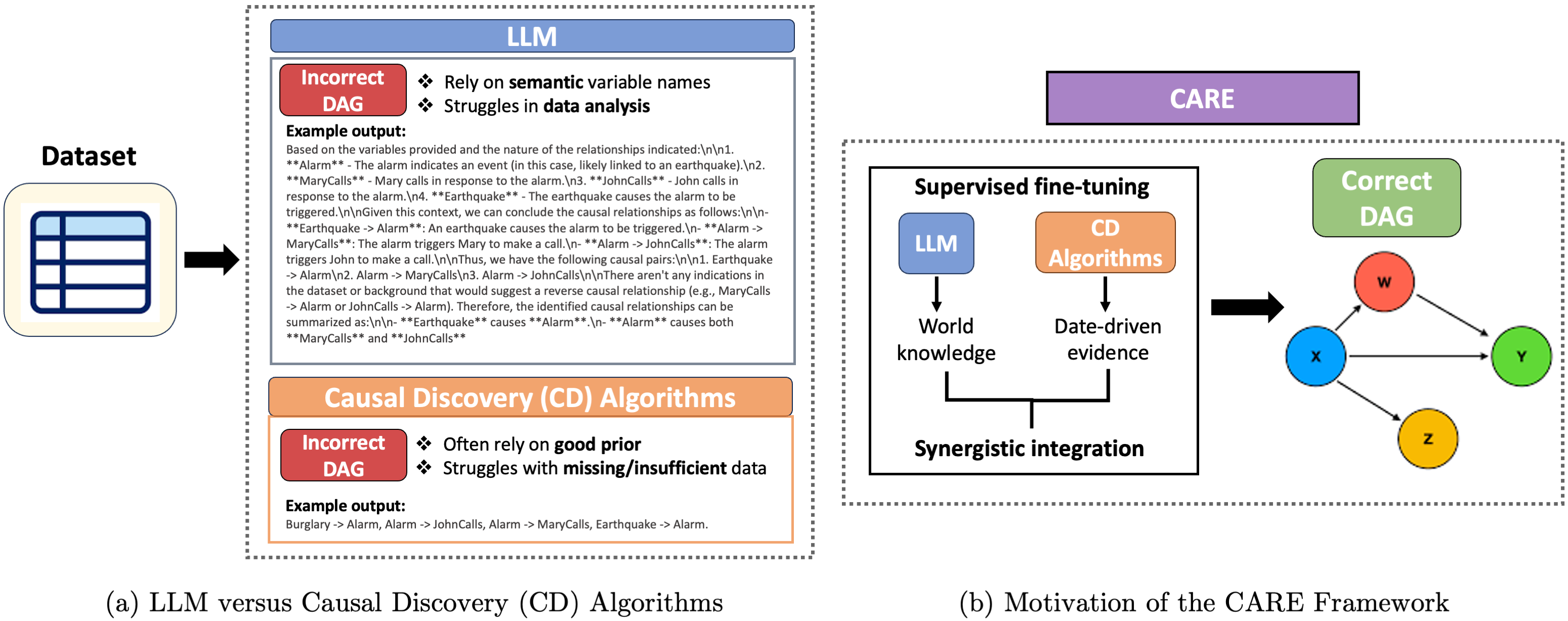}
  \vspace{-1em}
  \caption{\textbf{Synergizing LLM Knowledge with Strengths of Causal Discovery Algorithms}. {\bf (a)} Standalone approaches often falter: LLMs can misinterpret data due to over-reliance on pre-trained knowledge, while traditional Causal Discovery (CD) algorithms may struggle with insufficient data or without strong guiding priors, \emph{\color{Maroon}both potentially leading to incorrect causal graphs}. {\bf (b)} \CARE\ uses supervised fine-tuning to \emph{\color{Maroon}synergistically integrate} the extensive world knowledge of LLMs with the data-driven evidence from CD algorithm outputs, aiming to produce \emph{\color{Maroon}more accurate} causal discoveries.}
  \label{fig:teaster}
\end{figure}

This observation is perhaps not very surprising, given that LLMs are primarily trained for next-token prediction and natural language understanding, not for interpreting structured (e.g., tabular) data.
Yet, the ability to perform genuine causal discovery from data is \emph{crucial} for applying LLMs to domains where understanding cause-and-effect is essential. To address this challenge, our key insight is to prompt LLMs with the outputs of established causal discovery algorithms, which are \emph{inherently} designed for structured data, essentially treating these outputs as \emph{\color{Maroon}concise and sufficient summaries of the underlying statistical evidence} in a given observation dataset. Moreover, as shown in Figure~\ref{fig:teaster}, these algorithmic outputs are much more aligned with natural language representations, making them significantly easier for LLMs to interpret and utilize effectively. %\JD{See an example of prompting LLMs with causal discovery algorithm output to help causal discovery.}

However, quite surprisingly, we observe that LLMs cannot effectively use the algorithm output when they are injected as part of the prompts. Empirically, including algorithm outputs even sometimes \emph{\color{Maroon}degrades} the causal discovery performance of LLMs. To bridge these gaps, we propose \CARE\ (CAusal Reasoning Experts), a supervised fine-tuning (SFT) framework designed to empower LLMs to perform more robust and data-grounded causal structure learning. \CARE\ transforms pretrained LLMs by teaching them to \emph{\color{Maroon}synergistically integrate} their extensive {\bf world knowledge} with the {\bf structured outputs} of established causal discovery algorithms.
Our key technical contributions are twofold:

%To address these challenges, we propose \CARE, a supervised fine-tuning (SFT) framework to transform pretrained LLMs into \underline{CA}usal \underline{R}easoning \underline{E}xperts. 
%\CARE\; has multiple key technical contributions. 

%\textbf{\textit{(i)}} Rather than attempting to replace all the well-established causal inference methods with LLMs, the goal of \CARE\; is to teach LLMs to harness their strengths for improved causal reasoning. Our insight is that, while it is challenging for LLMs to effectively utilize a given dataset for causal queries, all the well-established causal inference algorithms were originally designed to complete this task. Thus, we consider the outputs from causal inference algorithms as sufficient statistics of the given dataset, and we use these algorithm outputs as inputs to the LLMs to help them better understand the information within the given dataset. 
%\textbf{\textit{(ii)}} Through a special token and SFT, we train LLMs to automatically select an appropriate causal inference algorithm for summarizing the dataset.  
%\textbf{\textit{(iii)}} We propose novel data augmentation methods which create challenging causal inference tasks where traditional causal inference methods fail. Through SFT on these tasks where outputs from causal inference methods do not lead to correct answers, LLMs learn to use its internal knowledge to calibrate the mistakes of causal inference methods. 

\begin{itemize}[left=5pt]

    \item[\ding{118}] \textbf{We investigate LLMs' capability in performing genuine causal discovery and their ability to employ algorithm outputs for self-improvement.} To this end, we consider multiple challenging scenarios  (detailed in Figure~\ref{fig:aug} and Section~\ref{subsec:augmenting_scenarios}) designed to test LLM biases, such as over-reliance on variable semantics and sensitivity to data ordering. 
    {\bf For example}, our benchmark includes a variable permuted version of the \textsl{EARTHQUAKE} dataset with $5$ variables including {\sc earthquake} and {\sc alarm}. While both the variables names and the statistical patterns within the observation dataset strongly indicate ``{\sc earthquake} causes {\sc alarm}'' in the original dataset, we create a different version of the earthquake dataset by exchanging the dataset columns corresponding to {\sc earthquake} and {\sc alarm}. By doing so, the statistical patterns within the dataset suggest that ``{\sc alarm} causes  {\sc earthquake}'' while the variable names suggest vice-versa. 
    
    \item[\ding{79}] \textbf{We propose a novel SFT paradigm for improved causal discovery.} \CARE\ is specifically designed to enhance the causal discovery performance of LLMs. Unlike approaches that rely solely on LLMs' internal knowledge or attempt direct data analysis, \CARE\ trains LLMs to effectively interpret and utilize the structured outputs (e.g., candidate graphs, conditional independencies) generated by established causal discovery algorithms. With \CARE, LLM learns to treat these algorithm outputs as sufficient evidence for data and correct these algorithms' biases with its own world knowledge. 
    {\bf For example}, when discovering a gene network, \CARE\ receives the graph proposed by an established causal discovery method based on expression data and learns to reason from that graph, potentially refining it using its world knowledge. This synergistic approach allows \CARE\ to outperform both baseline pretrained LLMs (which struggle with data-driven analysis) and often the standalone causal discovery algorithms themselves (which lack broader world knowledge for refinement), as demonstrated through our extensive experiments.

    % \item[\ding{60}] \textbf{We improve generalization and few-shot discovery.} We present evidence suggesting that \CARE\ enhances generalization capabilities on out-of-distribution or novel causal discovery tasks by learning to integrate data-derived evidence (via algorithm outputs) with their vast, pre-trained knowledge base. This synthesis may allow \CARE\ to make reasonable inferences even when faced with only a small number of data samples, situations where traditional algorithms often struggle due to insufficient statistical power. These results point towards a promising direction for applying LLMs to real-world discovery problems where data is often scarce.
\end{itemize}

\section{Related Work}
\label{sec:related_work}
Causal discovery aims to to learn causal relationships from observational and interventional data. There has been a wide array of algorithms that have been proposed, with different statistical assumptions and computational trade-offs. 

\textbf{Constraint-based methods.} Constraint-based algorithms try to recover the causal graph by exploiting a set of conditional independence tests. For example, the \texttt{PC} algorithm \citep{spirtes2000causation} is one of the first algorithm that has been proposed and it uses conditional independence tests to (e.g., Fisher's Z-test for Gaussian data or G-square for categorical data) to infer the skeleton of a causal graph and orient edges. %Fast Causal Inference (\texttt{FCI})~\citep{spirtes2000causation} is an extension of the \texttt{PC} algorithm that relaxes the assumption of no latent confounding and no selection bias in the observational setting.

\textbf{Score-based methods.} 
Score-based algorithms approach causal discovery by searching over the space of possible graphs to find the one that maximizes a predefined scoring function, which quantifies the goodness-of-fit between the graph structure \( G \) and the observed data~\citep{chickering2002optimal}. This typically involves balancing model complexity with data likelihood to avoid overfitting. Common scores are the Akaike Information Criterion (\texttt{AIC})~\citep{akaike2003new}, Bayesian Information Criterion (\texttt{BIC}), and the Bayesian Dirichlet sparse (\texttt{BDs})~\citep{scutari2016empirical}. Recent score-based methods include \texttt{BOSS} (Best Order Score Search)~\citep{andrews2023fast}, which leverages Grow-shrink trees~\citep{margaritis1999bayesian} for efficient exploration of the DAG space. %Greedy Relaxations of the Sparsest Permutation (\texttt{GRaSP}) ~\citep{lam2022greedy} utilizes permutation-based operation tuck, and they offer computationally efficient search of the DAG space under weaker assumptions than faithfulness.

\textbf{Functional methods.} These methods restricts the space of the Structural Causal Models function class. Notably, \texttt{ICA-LiNGAM}~\citep{shimizu2006linear} leverages the assumption of linear, non-Gaussian data generating processes to uniquely identify DAGs via independent component analysis. \texttt{DirectLiNGAM}~\citep{shimizu2011directlingam} extends this by sequentially identifying exogenous variables based on their statistical independence from residuals, enabling more scalable recovery of DAGs in practice.

\textbf{Causal Discovery and LLMs} The emergence of transformer-based architectures~\citep{vaswani2017attention} has led to a revolution in NLP and general AI capabilities. Modern LLMs, trained on massive text corpora, exhibit emergent reasoning capabilities and have shown promise in various symbolic and statistical reasoning tasks. Recent efforts have investigated whether LLMs can complement or even replace classical causal discovery tools. Some works explore LLMs’ capacity to understand causal relationships in language~\citep{ashwani2024cause, jin2023cladder}, revealing partial causal reasoning but limited structural understanding. Others aim to explicitly use LLMs to generate or assist with causal graphs~\citep{long2023can, jiang2023llm4causal}, bridging data-driven methods with prior knowledge encoded in language. Critically, debates remain as to whether LLMs can internalize or reason over structural causal models~\citep{chi2024unveiling}. 

While LLMs show promise in democratizing access to causal tools, their integration with traditional causal discovery algorithms remains an open frontier. Recent work~\citep{kiciman2024causal} categorizes the role of LLMs in causal discovery into three main paradigms: (a) direct inference from metadata, (b) posterior refinement of graphs produced by traditional algorithms, and (c) integration of LLM-derived priors into the discovery process. Within the direct inference paradigm, two primary approaches have emerged. The first focuses on causal order prediction using direct LLM queries combined with chain-of-thought prompting~\citep{willig2022can, kiciman2023causal, wei2022chain}. The second aims at full or partial causal graph discovery through iterative pairwise comparisons~\citep{long2023can}. However, this latter approach can incur substantial computational overhead, which has been addressed in more recent work on efficient inference strategies~\citep{jiralerspong2024efficient, sokolov2023towards}.
In the posterior correction literature, LLMs are used as expert judges to refine and correct the outputs of traditional causal discovery algorithms. Given the set of all conditional independence relations, ~\citep{long2023causal} proposes a method that treats the LLM as an imperfect expert to progressively narrow down the set of possible causal structures within the Markov equivalence class, while mitigating the risk of incorrect edge orientations. However, its scalability to larger datasets remains unproven, likely due to the method’s computational complexity and design limitations. LLMs can also be used to provide prior knowledge by extracting metadata from textual descriptions or domain-specific sources. One approach feeds variable descriptions to an LLM to infer direct causal relationships based on semantic understanding \citep{ban2023query}. Prior knowledge is typically incorporated using either hard or soft constraints. Hard constraints enforce strict exclusions of edges, but lack flexibility, errors in prior assumptions cannot be corrected during learning. To address this, some methods propose mechanisms for detecting and correcting flawed LLM-derived priors \citep{chen2023mitigating}.

% \begin{itemize}
%     \item ~\citep{spirtes2000causation} introduces the PC algorithm.
%     \item ~\citep{peters2017elements} a general causality book, it discusses the different causal discovery algorithms
%     \item 
% \end{itemize}

% \subsection{LLMs}

% \begin{itemize}
%     \item the transformer papaer~\citep{vaswani2017attention}
% \end{itemize}

% \subsection{LLMs and Causal Discovery}
% \begin{enumerate}
%     \item Cause and Effect: Can Large Language Models Truly Understand Causality? ~\citep{ashwani2024cause}
%     \item CLADDER: Assessing Causal Reasoning in
% Language Models ~\citep{jin2023cladder}
% \item LLM4Causal: Democratized Causal Tools for Everyone via Large Language Model ~\citep{jiang2023llm4causal}
% \item Unveiling causal reasoning in large language models: Reality or mirage? ~\citep{chi2024unveiling}
% \item Causal reasoning and large language models: Opening a new frontier for causality ~\citep{kiciman2024causal}
% \item Can Large Language Models Build Causal Graphs? - This paper evaluates if pretrained LLMs can be useful in complementing causal graph development~\citep{long2023can} 
% \item 
% \end{enumerate}
% }

\section{Preliminary}

\subsection{Causal Discovery}
In this section, we briefly review the causal discovery problem from a Structural Causal Model perspective ~\citep{pearl2009causality}. Let \( V \) denote a finite set of random variables, and \( G = (V, E) \) be a directed acyclic graph (DAG), where an edge \( (i, j) \in E \) encodes a direct causal influence from variable \( X_i \) to \( X_j \). Each node \( X_i \in V \) is generated according to a structural causal model (SCM):
\[
X_i = f_i(\mathrm{PA}_i, N_i),
\]
where \( \mathrm{PA}_i \subseteq V \setminus \{X_i\} \) denotes the parents of \( X_i \) in \( G \), and \( N_i \) is an exogenous noise variable independent of all other noise terms. Under standard assumptions, this implies a factorization,
\begin{equation}
\textstyle
\label{eq:markov}
P(X) = \prod_{i=1}^{|V|} P(X_i \mid \mathrm{PA}_i).
\end{equation}

The goal of causal discovery is to identify the true DAG, i.e., edges in $E$, from an observation dataset containing sampled according to the data generating process (DGP) specified above.
% Importantly, the true DAG structure is typically non-identifiable from only observation data. 
% Multiple DAGs can encode the same conditional independencies and yield the same distribution factorization. These DAGs form a \emph{Markov Equivalence Class} (MEC)~\citep{verma2022equivalence}. Two DAGs are said to be Markov equivalent if they entail the same set of conditional independence relations, which can be read off the graph using the criterion of \emph{d-separation}~\citep{pearl2009causality}.

%\begin{definition}[Markov Equivalence Class]
%Two DAGs \( G_1 \) and \( G_2 \) are Markov equivalent if for all disjoint subsets \( A, B, C \subseteq V \), we have
%\[
%A \indep B \mid C [G_1] \iff A \indep B \mid C [G_2].
%\]
%\end{definition}
%Traditional causal discovery methods often rely on two key assumptions. First, the causal Markov assumption: The distribution \( P \) satisfies the Markov property with respect to the true causal DAG \( G \): if \( A \indep B \mid C [G] \), then \( A \indep B \mid C [P] \). Second, the causal faithfulness assumption: The converse also holds: \( A \indep B \mid C [P] \Rightarrow A \indep B \mid C [G] \). This assumption rules out pathological parameterizations where conditional independencies cancel out direct causal effects.

\paragraph{Evaluation Metrics} 
To assess the quality of the learned causal graph, we treat each edge in the ground-truth set \( E \) as a target to be predicted, and compare it against the predicted edge set \( \hat{E} \). We choose the standard \textbf{F1 Score} as our main evaluation metrics, defined as
$$
    \text{F1} = \frac{2 \cdot \text{Precision} \cdot \text{Recall}}{\text{Precision} + \text{Recall}},
$$
where $
    \text{Precision} = \frac{|\hat{E} \cap E|}{|\hat{E}|},
    $ 
and 
$
    \text{Recall} = \frac{|\hat{E} \cap E|}{|E|}.
$

% \begin{itemize}[itemsep=0em, topsep=0em]
%     \item \textbf{Precision.} The proportion of predicted edges that are correct:
%     $
%     \text{Precision} = \frac{|\hat{E} \cap E|}{|\hat{E}|},
%     $
%     \item \textbf{Recall.} The proportion of true edges that are correctly predicted:
%     $
%     \text{Recall} = \frac{|\hat{E} \cap E|}{|E|}.
%     $
%     \item \textbf{F1 Score.} The harmonic mean of precision and recall:
%     $
%     \text{F1} = \frac{2 \cdot \text{Precision} \cdot \text{Recall}}{\text{Precision} + \text{Recall}}.
%     $
% \end{itemize}
    
    %\item \textbf{Structural Hamming Distance (SHD).} The number of edge additions, deletions, or reversals needed to convert the predicted graph \(\hat{G}\) into the ground-truth graph \(G\). Formally:
   % \[
    %\text{SHD}(\hat{G}, G) = |\hat{E} \setminus E| + |E \setminus \hat{E}| + \#\text{wrong orientations}.
    %\]

% \textbf{Metrics.} We measure:
% \begin{itemize}
% \item Method selection accuracy: Does the model pick the correct method family?
% \item Estimation error: Compare the estimated ATE/CATE with the true effect (RMSE or absolute error). \A{are we adding this? I thought we will just focus on causal discovery}
% \item Causal discovery accuracy: For generated graphs, we check if the model recovers the correct structure or directionality.
% \end{itemize}

\subsection{Large Language Models and Supervised Fine-Tuning}

LLMs are transformer-based architectures trained on large-scale text corpora using the objective of next-token prediction. Formally, given a sequence of tokens $x = (x_1, x_2, \dots, x_T)$ and model parameters $\theta$, the model is trained to maximize the likelihood,
\begin{equation}
\textstyle
\max_\theta \sum_{t=1}^{T} \log p_\theta(x_t \mid x_{<t}),
\end{equation}
where $x_{<t}$ represents the sequence prior to time step $t$. This autoregressive training objective enables LLMs to model complex dependencies in natural language and generate coherent text continuations.

Despite their impressive generalization capabilities, pretrained LLMs are often misaligned with task-specific goals or user preferences. To address this, a common practice is \emph{Supervised Fine-Tuning} (SFT), where a pretrained model is optimized on instruction-response pairs $\{(x^{(i)}, y^{(i)})\}_{i=1}^N$, where $x^{(i)}$ is typically a prompt or instruction, and $y^{(i)}$ is the desired output. The SFT objective is,
\begin{equation}
\textstyle
\max_\theta \sum_{i=1}^N \log p_\theta(y^{(i)} \mid x^{(i)}),
\end{equation}
SFT refines the model’s behavior to better follow human instructions, generate accurate task-specific outputs, and align with downstream usage scenarios. It plays a central role in adapting general-purpose LLMs to specialized domains such as reasoning, programming, and scientific discovery.

\section{Prompting LLMs with Causal Discovery Algorithm Outputs}
\label{sec:investigation}

\textbf{Motivation.} We begin by asking whether \emph{\color{Maroon}prompting alone} can boost LLMs' causal discovery capabilities.
Concretely, we study two questions:
\begin{enumerate}[itemsep=0em, topsep=0em]
    \item[\textbf{Q1.}] Does providing more observational data improve LLMs' ability to recover the underlying causal graph?
    \item[\textbf{Q2.}] Can LLMs, \emph{\color{Maroon}without} any additional finetuning, effectively leverage the outputs of established causal discovery algorithms when included in the prompt? 
    %If we include the outputs of established causal discovery algorithms into the prompts, can LLMs leverage this information to improve performance \emph{without} any additional finetuning?
\end{enumerate}

%\begin{enumerate}
%    \item[\textbf{Q1.}] Is it beneficial to present a larger observational dataset to the LLMs? 
%    \item[\textbf{Q2.}] Is it beneficial to include the outputs of established causal discovery algorithms into the prompts? 

 %   Specifically, we consider the following methods: PC (Fisher Z, G-Square), Greedy Equivalence Search (GES), ICA-LiNGAM, DirectLiNGAM, Fast Causal Inference (FCI), Greedy Sparsest Permutation (GRaSP), Bayesian Optimization Structure Search (BOSS), and the Hill-Climbing and PC algorithms as implemented in the BNlearn package.
%\end{enumerate}

In particular, answering \textbf{Q1} tells us whether LLMs can directly transform raw observational data into causal structure. Answering \textbf{Q2} tests their capacity to \emph{\color{Maroon}integrate and refine} external algorithmic knowledge. In particular, we are interested in whether LLMs can correct their and/or algorithmic biases to achieve a \emph{\color{Maroon}combined} effect where \emph{LLM+algorithm} outperforms either component alone. 
% Here, we conduct all experiments \emph{without task‑specific SFT} to isolate the benefits of prompting.

%In particular, the first question represents whether LLMs can effectively utilize the given observational dataset, and the second question aim to investigate whether LLMs can effectively use the outputs of established algorithms to improve its causal reasoning and decision \textit{without SFT}. 

\textbf{Experiments Setup.} We evaluate an array of LLMs: \textbf{(i)} \emph{Qwen2.5-1.5B}, \textbf{(ii)} \emph{gpt-4.1-mini}, \textbf{(iii)} \emph{gpt-4o-mini}, and \textbf{(iv)} \emph{o4-mini}, using the classic \emph{ASIA} benchmark with $8$ variables. 

To address \textbf{Q1}, we query the LLMs with prompts containing observational data samples drawn from the ASIA data distribution with increasing sample sizes $N \in \{0, 50, 200\}$. Performance is evaluated based on the F1-score comparing the LLM-generated causal graph against the ground truth. 

To address \textbf{Q2}, we first run a collection of established causal discovery algorithms with the observational data and collect their algorithm outputs including an estimated causal graph. We consider an assortment of representative algorithms: \texttt{PC}, \texttt{GES}, \texttt{ICA-LiNGAM}, \texttt{DirectLiNGAM}, \texttt{FCI}, \texttt{GRaSP}, \texttt{BOSS}. Details of these algorithms are discussed in Section~\ref{sec:related_work}.

For each sample size $N \in \{0, 50, 200\}$, we conduct a controlled comparison between two conditions: (a) LLMs prompted \emph{\color{Maroon}with} the algorithm outputs and (b) LLMs prompted \emph{\color{Maroon}without} them.

\textbf{Controlling for Prior Knowledge in Variable Semantics.} A critical consideration when evaluating LLMs’ ability to perform genuine causal analysis is their potential reliance on pre-existing knowledge of variable semantics~\citep{cai2023knowledge}. To isolate this factor and rigorously assess the models’ capacity for data-driven reasoning, we design experiments under three {\bf variable naming conditions}, each presenting a distinct challenge:
\begin{itemize}[left=5pt,itemsep=0em, topsep=0em]
\item \textbf{\emph{Original}} uses the standard, semantically meaningful names from the ASIA dataset, e.g., {\sc VisitAsia}, {\sc Smoking}, {\sc Tuberculosis}, etc. \emph{\color{Maroon}This allows LLMs to leverage their equipped world knowledge through variable name semantics}.
\item \textbf{\emph{Non-semantic (Random)}} replaces each variable name with a unique, non-semantic identifier consisting of randomly sampled characters (e.g., {\sc xray} $\rightarrow$ {\sc 4nws}). This ensures variable names carry no semantic meaning, \emph{\color{Maroon}forcing the LLMs to rely solely on the data structure and, if applicable, provided algorithm outputs}.
\item \textbf{\emph{Permuted}} shuffles variable names while leaving the underlying dataset matrix untouched; hence, we remove any reliable linguistic hints that a model might rely on. Moreover, this creates a challenging scenario for LLMs, as they need to \emph{\color{Maroon}counteract the biases associated with the semantic knowledge}. 
Success in this condition demands reasoning purely from the observation data while refraining from the potentially misleading variable names.
\end{itemize}

{\bf Results.} We summarize the key results in Table~\ref{tab:sec4}, with several important observations:
\begin{enumerate}[left=15pt,itemsep=0em, topsep=0em]
\item[{\bf O1.}] \textbf{Reliance on semantic priors.}
With original, semantically meaningful variable names in Table~\ref{tab:sec4}, LLMs often achieve \emph{\color{Maroon}high F1-scores}, \emph{\color{Maroon}even when with \textbf{no} observation data} nor algorithm outputs. 
% ,e.g., > 0.9 for GPT-4.1-mini, o4-mini, N=0 
% ,frequently outperforming causal algorithms 
% (algorithm performance: $\approx$ 0.36).
This suggests dominant reliance on memorized semantic associations rather than data-driven analysis.
\item[{\bf O2.}] \textbf{Performance collapse without semantic hints.}
When variable names are rendered non-informative—either via permuted variable names or non-semantic random names in Table~\ref{tab:sec4}, \emph{\color{Maroon}LLM performance plummets}, particularly when without algorithm outputs (often near zero F1-score). This highlights their inability to perform causal discovery from data patterns alone.

\item[{\bf O3.}] \textbf{Ineffective use of observational data (Q1).}
In semantically-neutral conditions (permuted variable names and non-semantic names), providing or \emph{\color{Maroon}increasing} observational data (N=0 to N=200) offers \emph{\color{Maroon}minimal and inconsistent improvement} -- LLMs do not effectively learn from raw data via prompting alone.

\item[{\bf O4.}] \textbf{Limited and non-synergistic use of algorithm outputs (Q2).}
While providing algorithm outputs improves performance over data-only prompting (without algorithm) in semantically-neutral settings, LLM performance typically only approaches or lags behind the input algorithm's standalone performance (e.g., Table~\ref{tab:sec4}, LLM scores mostly < 0.4 vs. Algorithm Performance 0.413). There's no clear evidence of LLMs refining or synergistically improving upon these outputs. With original names, algorithm outputs can even \emph{\color{Maroon}degrade} LLM performance if they conflict with LLMs' strong semantic priors.
\end{enumerate}

\begin{table}[H]
\centering
\caption{Performance with and without Algorithm Outputs (F-1 score).}
\scalebox{0.8}{
\begin{tabular}{llcccccccccc}
\toprule
\multirow{2}{*}{\textbf{Model}} & \multirow{2}{*}{\textbf{Conditions}} & \multicolumn{3}{c}{\textbf{Original}} & \multicolumn{3}{c}{\textbf{Random}} & \multicolumn{3}{c}{\textbf{Permuted}} \\
\cmidrule(lr){3-5} \cmidrule(lr){6-8} \cmidrule(lr){9-11}
 & & \textbf{N=0} & \textbf{N=50} & \textbf{N=200} &  \textbf{N=0} & \textbf{N=50} & \textbf{N=200} & \textbf{N=0} & \textbf{N=50} & \textbf{N=200} \\
\midrule
Algorithm Performance &      & --    & --     & 0.362   & --    & --     & 0.413    &  --   &  --    & 0.361     \\
\midrule
\multirow{2}{*}{\emph{Qwen2.5-1.5B}}
  & with Algo    &  0.609   &  0.496    & 0.423  & 0.327 &  0.328   & 0.344   &   0.357    &   0.300   &   0.338   \\
  & w/o Algo &   0.554  &  0.404    &  0.351  & 0.027 & 0.143   & 0.053   & 0.085    &  0.074    & 0.063     \\
\midrule
\multirow{2}{*}{\emph{GPT-4.1-mini}}
  & with Algo    &  0.812   &  0.858    &  0.840   & 0.314  &  0.315    & 0.314    &  0.132   &  0.137    &   0.127   \\
  & w/o Algo &  0.977   &  0.988    &  0.984  & 0.000  & 0.175   & 0.209    &  0.126   &   0.099   &   0.107   \\
\midrule
\multirow{2}{*}{\emph{GPT-4o-mini}}
  & with Algo    &  0.461   &   0.497   &   0.568   & 0.381  &  0.380  & 0.376   &  0.353   &  0.367    &  0.374    \\
  & w/o Algo &  0.688   &   0.693    & 0.637  &  0.000  & 0.129   & 0.132   &  0.133   &    0.136  &  0.125    \\
\midrule
\multirow{2}{*}{\emph{o4-mini}}
  & with Algo    &   0.986  &  0.994    &    0.994   & 0.293  &   0.219    & 0.271   &   0.164  &  0.152    &   0.152   \\
  & w/o Algo & 1.000    &  0.996    &  1.000    & 0.000  & 0.078   & 0.052  &  0.160   &   0.149   &    0.153   \\
\bottomrule
\end{tabular}
}
\label{tab:sec4} 
\end{table}

\textbf{Summary.} This exploratory investigation suggests that prompting LLMs for causal discovery is highly unreliable when semantic meanings are unavailable. LLMs primarily use variable name semantics, struggle to analyze observational data effectively, and do not consistently or synergistically enhance outputs from established causal discovery algorithms. The absence of robust data-driven reasoning and critical engagement with algorithmic outputs highlight \emph{\color{Maroon}the inadequacy of prompting alone for this complex task}. These findings motivates us to use SFT to teach LLMs the specific skills required for genuine causal discovery from diverse information sources.

\section{\CARE: Transforming LLMs into Causal Reasoning Experts}
\label{sec:care_framework}

\begin{figure}[th]
  \centering
  \includegraphics[width=\linewidth]{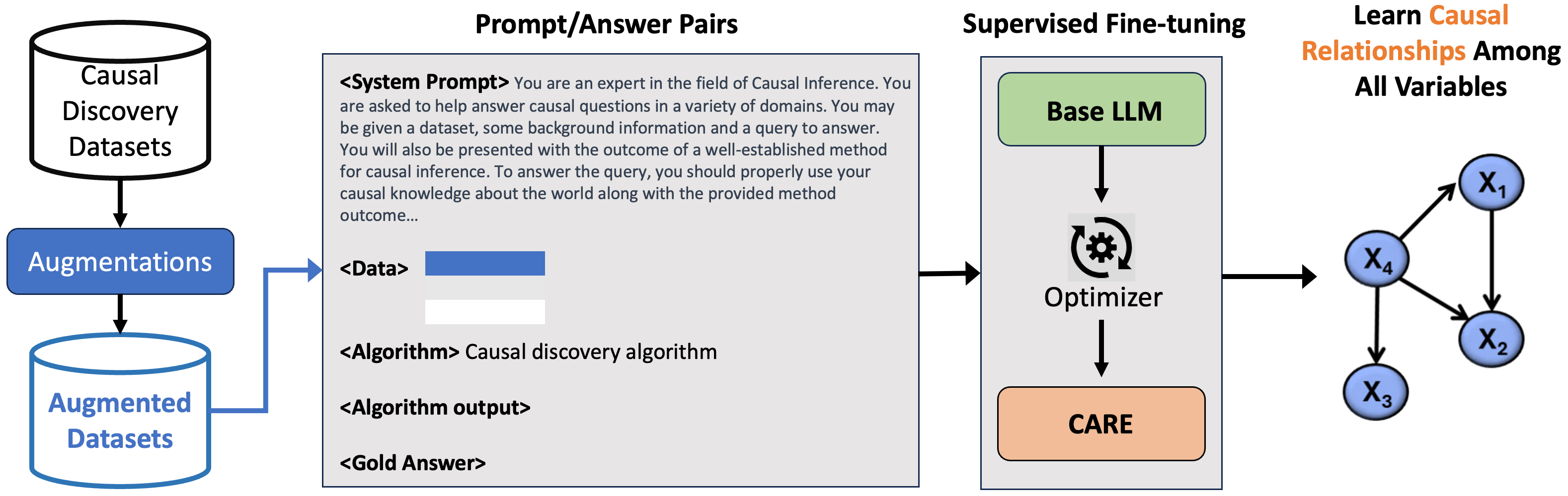}
  %\vspace{-1em}
  \caption{\textbf{\CARE\ Framework Overview.} \CARE\ processes causal discovery datasets, augments them to create \emph{\color{Maroon}diverse training scenarios} ({\bf left}), and uses these to construct \emph{\color{Maroon}prompt/answer pairs} ({\bf middle-left}). These pairs are then used for \emph{\color{Maroon}supervised fine-tuning} ({\bf middle-right}), enabling LLMs to learn and output accurate causal relationships among variables ({\bf right}).}
  \label{fig:flowchart}
\end{figure}

Given the limitations of prompting identified in Section~\ref{sec:investigation}, particularly the surprising finding that providing outputs from strong causal discovery algorithms can decrease LLM performance, we aim for an approach to transform LLMs into robust causal reasoning experts via SFT. To this end, we propose \CARE~(CAusal Reasoning Experts). See Figure~\ref{fig:flowchart} for the workflow of our framework.

\subsection{Framework Overview}
\textbf{Core Objective.} \CARE~trains an LLM to effectively combine its internal semantic knowledge with external algorithm outputs, specifically for the purpose of discovering more accurate causal graphs. Specifically, we leverage SFT to teach LLMs how to synthesize these diverse information sources while ensuring resilience against data variations.
% The LLM learns to combine its latent understanding of variable relationships (often derived from semantics) with explicit structural hypotheses generated by causal discovery algorithms run on observational data.

\textbf{Training objective.} The LLM is trained to output a \emph{\color{Maroon}refined causal graph structure} that represents the optimal synthesis of the LLM's internal priors and external algorithmic evidence. 
% The target is to better approximate the true underlying causal graph.

\textbf{SFT Instances.} Our SFT procedure follows the standard instruction-following template with each SFT instance consisting of an \emph{\color{Maroon}instruction} and a \emph{\color{Maroon}target}. 
The {\bf instruction} includes {\bf (a)} the context or query defining the discovery task, e.g., ``Find the causal graph for variables {\sc V1}, {\sc V2},...''; {\bf (b)} an observational dataset; and {\bf (c)} the output graph structures from one or more established causal discovery algorithms (e.g., \texttt{PC}, \texttt{GES}) applied to the dataset. 
% \begin{itemize}
% \item[(a)] The context or query defining the discovery task (e.g., "Find the causal graph for variables V1, V2,...").
% \item[(b)] The observational dataset.
% \item[(c)] The \textbf{output graph structures from one or more established causal discovery algorithms} (e.g., PC, GES,LiNGAM) applied to the dataset (b). These algorithm outputs serve as the primary structured, data-driven evidence.
% \end{itemize}
The {\bf target} includes the ground-truth causal graph corresponding to the synthetic data scenario. 
By training against this target, LLMs autonomously learn \emph{\color{Maroon}the optimal strategy for weighting and integrating} the semantic context and the provided algorithm outputs to best recover the true structure.
% \paragraph{Target Formulation:} The desired output during SFT is the \textbf{ground-truth causal graph} corresponding to the synthetic data scenario. By training against this target, the LLM implicitly learns the optimal strategy for weighting and integrating the semantic context and the provided algorithm outputs to best recover the true structure.
We next elaborate on our procedure for generating the SFT data, a central challenge to ensure that LLMs generalize.

\subsection{Data Augmentation Strategy for Robust Learning.}
\label{subsec:augmenting_scenarios}
%Robust causal discovery should rely on \emph{invariant statistical structure} rather than superficial quirks of how data are named, ordered, or partially observed. We therefore use a collection of complementary augmentations (as shown in Fig~\ref{fig:aug}), each designed to stress‑test a specific weakness that naive models often exploit:

\textbf{Motivation.} The findings in Section~\ref{sec:investigation} demonstrate that \emph{\color{Maroon}prompting alone is insufficient} to elicit robust, data-driven causal discovery from LLMs. Specifically, LLMs tend to over-rely on variable semantics and struggle to effectively integrate observational data or external algorithmic outputs. To transcend these limitations, SFT within our \CARE\ framework requires a training set that actively \emph{\color{Maroon}challenges these ingrained biases and teaches genuine evidence integration}. 
In particular, an effective SFT requires a large and diverse dataset of causal discovery problems where combining internal knowledge and external algorithm outputs is beneficial. 
To this end, we employ a collection of carefully designed data augmentations.

\begin{figure}[h]
  \centering
  \includegraphics[width=0.8\linewidth]{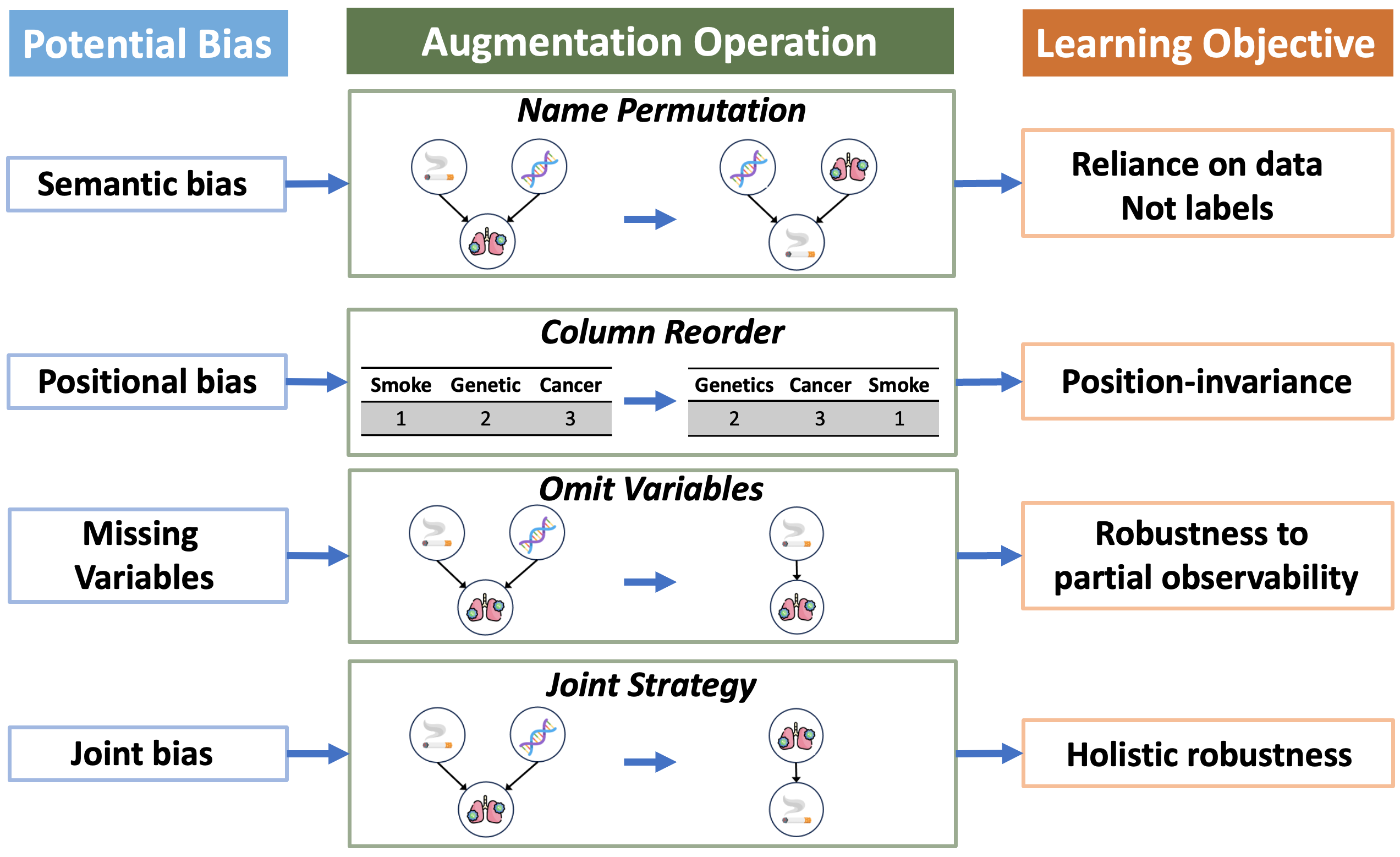}
  %\vspace{-1em}
  \caption{\textbf{Purposes of Various Augmentation Methods.}
  Each row illustrates a potential LLM \emph{\color{Maroon}bias} ({\bf left}), the corresponding \emph{\color{Maroon}augmentation} operation designed to address it ({\bf middle}), and the intended \emph{\color{Maroon}learning objective} for the model ({\bf right}).}
  \label{fig:aug}
\end{figure}

Concretely, our SFT data generation strategy is designed to ensure that the LLMs learn to:
\begin{enumerate}[itemsep=0em, topsep=0em]
\item[\textbf{A1.}] Reason from statistical structure \textit{independent of misleading or absent} semantic cues.
\item[\textbf{A2.}] Maintain performance irrespective of superficial data presentation characteristics like \textit{column ordering}.
\item[\textbf{A3.}] Perform robust discovery even with \textit{incomplete information} due to unobserved variables.
\item[\textbf{A4.}] Synergistically combine its internal knowledge with external algorithmic evidence, especially under \textit{complex, multifaceted perturbations}.
\end{enumerate}
To achieve (\textbf{A1-A4}), we rely on various  data augmentation methods for SFT (see Figure~\ref{fig:aug}). Each augmentation is designed for a specific bias, fostering \emph{\color{Maroon}more generalizable} causal reasoning. Specifically, we use four types of augmentation, one for each of {\color{Maroon}\textbf{A1-A4}}:
%While the original dataset provide the core learning material, fine-tuning solely on their original form might lead the LLM to learn superficial correlations or memorize specific patterns rather than acquiring a generalizable skill. To ensure robustness and force the model to learn the underlying principles of evidence integration for causal discovery, we apply a collection targeted data augmentations during the SFT process (summarized in Figure~\ref{fig:aug}).
\begin{itemize}[left=5pt,itemsep=0em, topsep=0em]
\item \textbf{Semantic Variable Name Permutation {\color{Maroon}[A1]}:} Original semantic variable names are shuffled; data matrix remain unchanged.
\item \textbf{Column Order Randomization {\color{Maroon}[A2]}:} Variable column sequence is permuted.
\item \textbf{Variable Omission {\color{Maroon}[A3]}:} Remove subset of the variables and update the ground-truth causal graph to reflect the marginalized structure over the remaining variables. 
\item \textbf{Perturbation Combination {\color{Maroon}[A4]}:} Compositions of above to ensure holistic robustness. This collection evaluates robustness using A2 as a foundational perturbation, either alone or in combination with other perturbations. Based on \emph{\color{Maroon}how the perturbations affect ground truth}, the specific scenarios are (i) \textit{Original:} the foundational scenario applying only perturbation A2; (ii) \textit{Omitted:} combining perturbation A3 with the foundational A2; (iii) \textit{Permuted:} combining perturbation A1 with the foundational A2. 
% \begin{itemize}[left=5pt,itemsep=0.5pt]
%  \item \textbf{Original:} The foundational scenario applying only perturbation R2.
%   \item \textbf{Omitted:} Combines perturbation R3 with the foundational R2.
%  \item \textbf{Permuted:} Combines perturbation R1 with the foundational R2. 
%  \end{itemize}
\end{itemize}

By training \CARE\ across this diverse corpus with thousands of augmented datasets derived from each benchmark dataset, we teach the model to learn \emph{\color{Maroon}robust and generalizable causal knowledge}. 

%In other words, we teach the model to learn the conditional independence relationships dictated by the underlying graph structure, which remain invariant despite transformations in naming, ordering, or observability. This rigorous approach moves beyond surface-level pattern matching towards identifying robust and generalizable causal knowledge.

\textbf{A Bayesian Perspective on SFT.} We view the pretrained LLM’s background knowledge as a \emph{prior} over causal structures: it encodes statistical and semantic associations learned from large-scale text corpora, which can implicitly inform causal relationships. During SFT, this prior is updated into a \emph{posterior} by incorporating new evidence, namely, observational data and the outputs of classical causal discovery algorithms. In this way, the LLM learns to integrate its internal knowledge with relevant data-driven clues, resulting in more accurate and grounded causal discovery predictions.

\section{Experiments}
\label{sec:experiments}
%\subsection{Datasets}

We empirically evaluate the effectiveness of our \CARE\ framework in enhancing the causal discovery capabilities of LLMs through Supervised Fine-Tuning (SFT). Our fine-tuned model is benchmarked against strong baseline LLMs and the performance of traditional causal discovery algorithms.

\textbf{Training Mechanism.} To manage computational resources, we utilize Parameter-Efficient Fine-Tuning (PEFT), specifically Low-Rank Adaptation (LoRA) \citep{hu2022lora}, enabling efficient adaptation of large base models by training only a small subset of their parameters. In addition, PEFT can help mitigate overfitting and reduce the risk of catastrophic forgetting in LLMs.

\subsection{Evaluation Strategy}
\label{eval_strategy} 

To evaluate the performance of our fine-tuned models, we employ an LLM-as-a-judge approach, leveraging a capable model for automated assessment. Here, we use \texttt{GPT-4.1-mini}. The evaluation process for each test instance is as follows::
\begin{enumerate}[left=5pt,itemsep=0em, topsep=0em]
    \item \textbf{Structured comparison:} The evaluation LLM receives the ground-truth causal graph and the graph predicted by the model under test. It is instructed via a detailed prompt (see Appendix \ref{suppsub:eval}) to parse directed edges from both, handling formatting variations and prioritizing specifically marked answer sections in model outputs.
    \item \textbf{Edge metrics calculation:} The judge LLM identifies and counts true positives (TP, correctly recovered edges), false negatives (FN, missed true edges), and false positives (FP, incorrectly predicted extra edges).
    \item \textbf{Performance metric:} From the TP, FP, and FN counts, we calculate the F1-score, a standard metric for graph discovery accuracy.
\end{enumerate}

This automated pipeline ensures consistent evaluation across different models, datasets, and augmentation types.

\subsection{Experimental Setup}

\textbf{Datasets.} Our experiments utilize established benchmark networks from the \texttt{bnlearn} repository\footnote{\url{https://www.bnlearn.com/bnrepository/}}. These include: \textsl{ASIA} (8 nodes)\citep{lauritzen1988local},  \textsl{SURVEY} (6 nodes)\citep{scutari2021bayesian}, \textsl{EARTHQUAKE} (5 nodes)\citep{korb2010bayesian}, and the larger \textsl{ALARM} network (37 nodes)\citep{beinlich1989alarm}. Each network provides a known ground-truth Directed Acyclic Graph (DAG), allowing for precise quantitative assessment. Further details on these datasets are in Appendix~\ref{supp:datasets}.

Synthetic observational data is sampled from the true conditional probability distributions of these networks. This data serves as the foundation for our augmentation strategy (detailed in Section~\ref{subsec:augmenting_scenarios} and Figure~\ref{fig:aug}), which applies Name Permutation, Column Reordering, Variable Omission, and Joint Perturbations to create a diverse and challenging corpus for robust training and testing.

A crucial aspect of our evaluation methodology is that for all test scenarios, we generate \textbf{\emph{new, unseen observational datasets}} sampled from the ground-truth graphs. While in-distribution evaluation utilizes graph structures encountered during SFT, testing on these distinct data realizations ensures we assess genuine generalization to novel data instances rather than mere memorization of the training samples. This significantly increases the credibility of our assessment.

\textbf{Benchmark Models.} We evaluate the following models:
\begin{itemize}[left=5pt,itemsep=0em, topsep=0em]
\item \textbf{Baseline LLMs:} We directly prompt the base \texttt{Qwen2.5-1.5B}, \texttt{gpt-4.1-mini}, \texttt{gpt-4o-mini}, \texttt{o4-mini} with algorithm outputs \emph{\color{Maroon} with no additional fine-tuning}. 
% These are prompted with algorithm outputs \emph{\color{Maroon} with no additional fine-tuning}.
\item \textbf{Algorithm:} The F1-score achieved by the collection of traditional causal discovery algorithms whose outputs serve as input to the LLMs. This acts as a non-LLM baseline.
\item \textbf{\CARE:} Our \emph{\color{Maroon}fine-tuned} \texttt{Qwen2.5-1.5B} model, using the SFT strategy described in Section~\ref{sec:care_framework}.
\end{itemize}

\subsection{Results and Analysis}

The main results comparing our fine-tuned \texttt{Qwen2.5-1.5B} model (\CARE) against baselines are presented in Table~\ref{tab:perf_in_distribution}. We analyze performance across the four benchmark datasets under three augmentation conditions: Original variable names, Omitted variables, and Permuted variable names as detailed in Section~\ref{subsec:augmenting_scenarios}. We demonstrate that our framework \CARE\ achieves state-of-the-art performance compared to baselines.

\begin{table}[H]
\centering
\caption{F1 score on benchmark datasets. Higher is better. Best results per condition are bolded.}
\label{tab:perf_in_distribution}
\resizebox{\linewidth}{!}{%
\begin{tabular}{l
                *{3}{c}  % Asia
                *{3}{c}  % Survey
                *{3}{c}  % Earthquake
                *{3}{c}} % Alarm
\toprule
\multicolumn{1}{c}{} &
\multicolumn{3}{c}{\textbf{Asia}} &
\multicolumn{3}{c}{\textbf{Survey}} &
\multicolumn{3}{c}{\textbf{Earthquake}} &
\multicolumn{3}{c}{\textbf{Alarm}} \\
\cmidrule(lr){2-4}\cmidrule(lr){5-7}\cmidrule(lr){8-10}\cmidrule(lr){11-13}
\textbf{Model} &
Original\ & Omitted\ & Permuted\, &
Original\ & Omitted\ & Permuted\, &
Original\ & Omitted\ & Permuted\, &
Original\ & Omitted\ & Permuted\, \\
\midrule
Algorithm           & 0.362 & 0.371 & 0.361 & 0.282 & 0.299 & 0.271 & 0.419 & 0.481 & 0.385 & 0.301 & 0.314 & 0.326 \\
\emph{GPT‑4.1‑mini}             & 0.858 & 0.824 & 0.137 & 0.268 & 0.269 & 0.269 & \textbf{1.000} & 0.979 & 0.220 & 0.425 & 0.395 & 0.446 \\
\emph{GPT‑4o‑mini}              & 0.497 & 0.380 & 0.367 & 0.313 & 0.308 & 0.293 & 0.447 & 0.549 & 0.437 & 0.420 & 0.458 & 0.529 \\
\emph{o4‑mini}                  & 0.994 & 0.978 & 0.152 & 0.319 & 0.301 & 0.246 & \textbf{1.000} & 0.971 & 0.242 & 0.487 & 0.507 & 0.415 \\
\emph{Qwen-2.5‑1.5B} & 0.496 &  0.336 & 0.300  & 0.306 & 0.293 & 0.268   & 0.397 & 0.475 & 0.404  & 0.400 & 0.362 & 0.409\\
\midrule
\CARE\ & \textbf{1.000} & \textbf{0.991} & \textbf{0.460} &
\textbf{0.979} & \textbf{0.997} & \textbf{0.792} &
0.953 & \textbf{1.000} & \textbf{0.623} &
\textbf{0.990} & \textbf{0.971} & \textbf{0.618} \\
\bottomrule
\end{tabular}
}
\end{table}

\section{Conclusion}

We introduced \CARE, an SFT framework designed to overcome LLMs' inherent limitations in causal discovery. By fine-tuning with diverse data augmentations that challenge semantic biases and promote data-driven analysis, \CARE\ enables LLMs to effectively synthesize their world knowledge with outputs from established causal algorithms. Our results demonstrate that a \CARE-finetuned Qwen2.5-1.5B model achieves state-of-the-art performance, surpassing both traditional methods and larger LLMs on challenging causal discovery tasks, particularly when variable names are permuted or information is partial. This highlights the efficacy of targeted SFT in building more capable and reliable causal reasoning agents from LLMs.

\bibliography{ref}
\bibliographystyle{abbrvnat}

\newpage

\appendix

\section{Supervised Fine-tuning}
\label{supp:SFT}

Here we discuss the Supervised Fine-tuning (SFT) used within the \CARE\ framework. As established in Section~\ref{sec:investigation}, prompting LLMs directly for causal discovery tasks, even with the inclusion of observational data or outputs from established causal discovery algorithms, yields unreliable performance and does not effectively address their inherent biases (e.g., over-reliance on variable name semantics). SFT within \CARE\ is therefore crucial for explicitly teaching the LLM to:
\begin{itemize}
    \item Robustly reason about causal structures from diverse data presentations.
    \item Synergistically integrate its internal world knowledge with the evidence provided by external causal discovery algorithms.
    \item Mitigate common biases observed in pre-trained models when faced with causal tasks.
\end{itemize}

The SFT process involves data preparation by using the augmentation strategies detailed in Section~\ref{subsec:augmenting_scenarios}, to structure causal problems into a conversational format. This is then followed by model training using established fine-tuning approaches.

\subsection{System Prompts for SFT}
\label{supp:sft_system_prompts}

Here we use two distinct system prompts to set the context and expected behavior for the model during SFT. The choice between these prompts is determined by whether an external causal discovery method's output is provided as part of the input for a given training instance. These prompts guide the LLM on how to approach the causal discovery task, emphasizing the use of provided information.

\begin{tcolorbox}[colback=Bittersweet!5!white,colframe=Bittersweet!80!black,title=System Prompt (with algorithm output)]
You are an expert in the field of Causal Inference. You are asked to help answer causal questions in a variety of domains. You may be given a dataset, some background information and a query to answer. You will also be presented with the outcome of a well-established method for causal inference. To answer the query, you should properly use your causal knowledge about the world along with the provided method outcome. Note the outcome of the causal inference method may not always be correct. Your goal is to provide the correct answer to the query. Do NOT invent new variables. If you invent new variables, your answer will be WRONG.
\end{tcolorbox}

\begin{tcolorbox}[colback=MidnightBlue!5!white,colframe=MidnightBlue!75!black,title=System Prompt (without algorithm output)]
You are an expert in the field of Causal Inference. You are asked to help answer causal questions in a variety of domains. You may be given a dataset, some background information and a query to answer. To answer the query, you should properly use your causal knowledge about the world along with the provided dataset if applicable. Your goal is to provide the correct answer to the query. Do NOT invent new variables. If you invent new variables, your answer will be WRONG.
\end{tcolorbox}

\subsection{Data Preparation for Fine-tuning}
\label{supp:sft_data_prep}

The foundation of SFT lies in generating high-quality training examples that mimic a dialogue format. Each example consists of a system directive (as detailed in Section~\ref{supp:sft_system_prompts}), a user query presenting a causal problem, and an ideal assistant response.

\subsubsection{User Message Construction: Presenting the Causal Problem}
User messages are formulated to present the causal problem to the model. This involves assembling various pieces of information from the source data:
\begin{itemize}
    \item \textbf{Background Information and Query}: The core context and the specific causal question to be answered.
    \item \textbf{Dataset (Optional)}: If applicable, variable names are extracted, and a specified number of sample data rows from the dataset can be included.
    \item \textbf{Causal Discovery Method Output (Optional)}: If provided in the source data, the output from a causal discovery algorithm (typically a graph structure) is included, after stripping away any performance metrics or extra details.
\end{itemize}

We want to highlight that important metadata fields like \texttt{`Correct Answer`} or \texttt{`Augmentation Type`} are \textbf{intentionally excluded} from the user prompt.

\subsubsection{Assistant Message Construction: Defining the Target Output}
The assistant's message in each training example represents the desired output. This is constructed by taking the pre-defined correct answer for the causal problem and formatting it, typically by prefixing it with a label like \texttt{***Answer***}.

\subsubsection{Tokenization and Labeling Strategy}
Once the conversational messages (system, user, assistant) are constructed, they are tokenized. This process converts the text into a sequence of numerical IDs that the language model can process. A standard chat template is applied to format the entire conversation into a single input sequence.

For training, two key sequences are generated:
\begin{itemize}
    \item \texttt{input\_ids}: The tokenized representation of the full conversation.
    \item \texttt{labels}: These are initially a copy of the \texttt{input\_ids}. A crucial step here is the option for ``answer-only'' training. If this mode is active, the parts of the \texttt{labels} corresponding to the system prompt and the user message (including the assistant's role identifier before its actual content) are masked (typically by setting their values to -100). This masking ensures that the model's learning (loss calculation) is focused solely on predicting the tokens of the assistant's answer, rather than the prompt itself.
\end{itemize}

\subsubsection{Dataset Handling}
Each data instance is then processed through the message construction and tokenization steps described previously, transforming it into the required \texttt{input\_ids} and \texttt{labels} format for the trainer. The \textbf{number of data samples} included in the \texttt{Dataset} field in the user prompt is configurable.

\subsubsection{Model and Tokenizer Setup}
\begin{itemize}
    \item \textbf{Tokenizer}: A tokenizer corresponding to a specified base model is loaded. It is configured with the end-of-sequence (EOS) token also serving as the padding token, and padding is applied to the left side of sequences. This is common for decoder-only models to ensure correct attention masking.
    \item \textbf{Model Loading}: The core causal language model is loaded from a model identifier.
    \begin{itemize}
        \item \textbf{Quantization (Optional)}: To manage memory and potentially improve efficiency, an option exists to load the model with 4-bit quantization. This uses techniques like NF4 quantization and \texttt{bfloat16} for computation, which can be particularly beneficial for training larger models on hardware with limited memory.
        \item If quantization is not enabled, the model is loaded in its standard precision.
    \end{itemize}
\end{itemize}

\subsubsection{Fine-tuning Strategies}
Here we use \textbf{Parameter-Efficient Fine-Tuning (PEFT)} with Low-Rank Adaptation (LoRA) \citep{hu2022lora}. Specifically LoRA enables efficient adaptation of large base models by training only a small subset of their parameters. In addition, PEFT can help mitigate overfitting and reduce the risk of catastrophic forgetting in LLMs.

\subsubsection{Trainer Configuration and Execution}
A specialized trainer, such as the \texttt{SFTTrainer} from the TRL library, is configured with various hyperparameters. These include:
\begin{itemize}
    \item Batch size per device and gradient accumulation steps (which together determine the effective batch size).
    \item Learning rate.
    \item Number of training epochs.
    \item Maximum sequence length to handle potentially long inputs.
    \item Settings for mixed-precision training (e.g., using FP16 or BF16) to accelerate training and reduce memory usage.
\end{itemize}
The trainer is then initialized with the prepared model (potentially adapted with LoRA), the tokenized dataset, and the training configuration. The training loop is initiated, and upon completion, the final trained model (or the LoRA adapter weights if PEFT was used) is saved.

\subsection{SFT Configuration Used}
\label{supp:sft_config_used}

For our experiments, the Supervised Fine-tuning (SFT) process utilized the \texttt{Qwen2.5-1.5B-Instruct} model as both the base for fine-tuning and for its corresponding tokenizer.

The model was trained for 30 epochs. A key aspect of this configuration was the use of a very long maximum sequence length, set to 15,000 tokens, enabling the model to process extensive contextual information within each training example. Furthermore, Parameter-Efficient Fine-Tuning (PEFT) was employed using Low-Rank Adaptation (LoRA) to optimize the training process by only updating a small subset of the model's parameters.

Other SFT training parameters, such as the learning rate, per-device batch size, gradient accumulation steps, and specific LoRA configurations (e.g., LoRA rank \texttt{r} and \texttt{lora\_alpha}), were set to their default values as defined within the SFT script and the underlying TRL library configurations. For instance, the LoRA rank (\texttt{r}) was 8, and \texttt{lora\_alpha} was 16. Training was performed with a batch size of 7 per device and 4 gradient accumulation steps. 

\section{Datasets}
\label{supp:datasets}

\begin{figure}[h] 
    \centering % Center the whole figure environment content
    \begin{minipage}[b]{0.48\textwidth} 
        \centering
        \includegraphics[width=\linewidth]{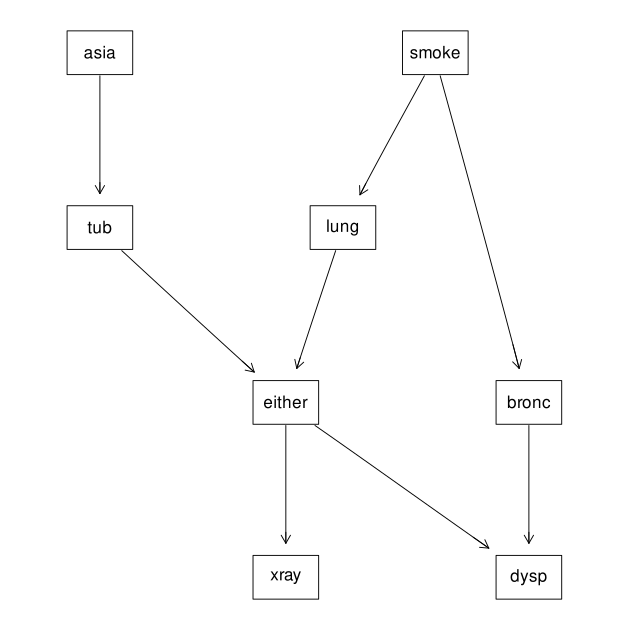} % \linewidth refers to width of minipage
        \caption{Ground-truth DAG for the \textsl{ASIA} network (8 nodes) \citep{lauritzen1988local}.}
        \label{fig:dag_asia}
    \end{minipage}
    \hfill
    \begin{minipage}[b]{0.48\textwidth} 
        \centering
        \includegraphics[width=\linewidth]{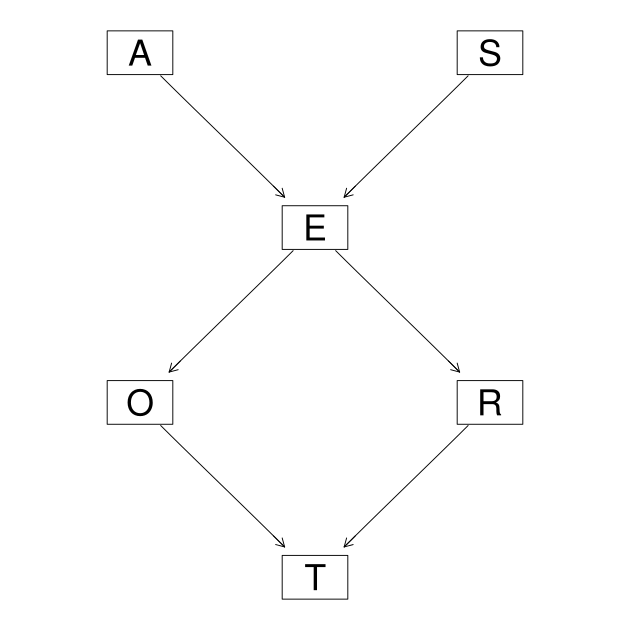} %
        \caption{Ground-truth DAG for the \textsl{SURVEY} network (6 nodes) \citep{scutari2021bayesian}.}
        \label{fig:dag_survey}
    \end{minipage}
    \caption{Ground-truth DAGs for the \textsl{ASIA} and \textsl{SURVEY} benchmark networks.} 
    \label{fig:dag_asia_survey} 
\end{figure}

\begin{figure}[h]
    \centering
    \includegraphics[width=0.6\textwidth]{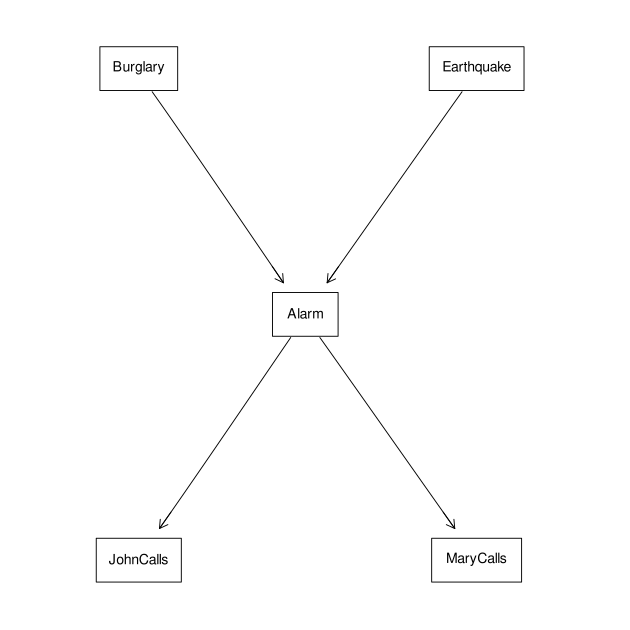} % Assuming earthquake.pdf exists
    \caption{Ground-truth DAG for the \textsl{EARTHQUAKE} network (5 nodes) \citep{korb2010bayesian}.}
    \label{fig:dag_earthquake}
\end{figure}

\begin{figure}[h]
    \centering
    \includegraphics[width=1.0\textwidth]{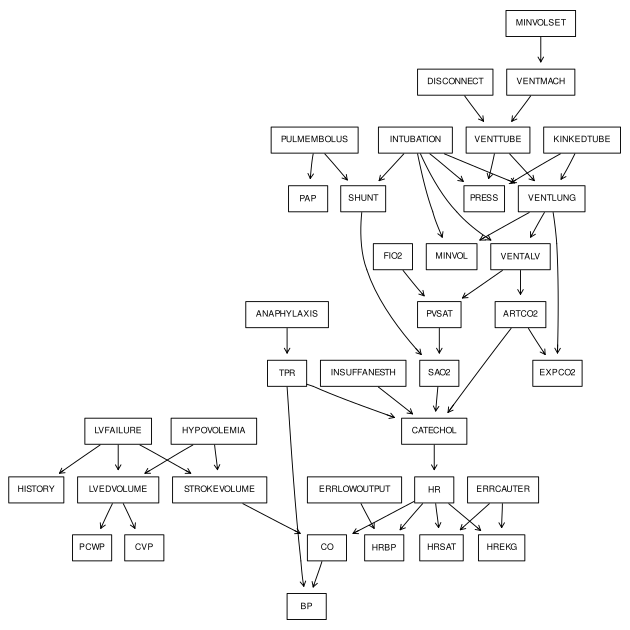} % Assuming alarm.pdf exists; might need a larger width
    \caption{Ground-truth DAG for the \textsl{ALARM} network (37 nodes) \citep{beinlich1989alarm}. A medical diagnostic network designed for patient monitoring, representing various physiological states, diseases, and measurements.}
    \label{fig:dag_alarm}
\end{figure}

The experiments conducted to evaluate the \CARE\ framework utilize several well-established benchmark Bayesian networks. These networks provide known ground-truth Directed Acyclic Graphs (DAGs), which are essential for quantitatively assessing the performance of causal discovery methods. The primary source for these networks is the \texttt{bnlearn} repository\footnote{\url{https://www.bnlearn.com/bnrepository/}}. The ground-truth DAG structures for these benchmarks are illustrated in Figures~\ref{fig:dag_asia} through \ref{fig:dag_alarm}.

The benchmark datasets used are:
\begin{itemize}
    \item \textbf{\textsl{ASIA}}: A small Bayesian network with 8 nodes, originally proposed by \citet{lauritzen1988local}. It represents a stylized medical diagnostic scenario involving variables such as recent travel to Asia, smoking, and respiratory diseases. The network models dependencies among factors like tuberculosis, lung cancer, bronchitis, and symptoms such as dyspnoea and X-ray results.
    \item \textbf{\textsl{SURVEY}}: A network with 6 nodes, described by \citet{scutari2021bayesian}. It models socioeconomic and demographic factors, including age (A), sex (S), education (E), occupation (O), residence (R), and preferred mode of transport (T). All variables are categorical.
    \item \textbf{\textsl{EARTHQUAKE}}: A compact network with 5 nodes, often attributed to \citet{korb2010bayesian}. It models a burglary direction scenario where an alarm may be triggered either by a burglary or an earthquake, and two neighbors (John and Mary) may or may not call the police based on hearing the alarm, each with a different reliability. 
    \item \textbf{\textsl{ALARM}}: A significantly larger network with 37 nodes, developed by \citet{beinlich1989alarm}. It was designed as a diagnostic support system for patient monitoring in intensive care, encoding relationships among physiological variables, diseases, and treatments. The network includes 37 nodes spanning diagnoses, findings, and intermediate variables to facilitate probabilistic reasoning under uncertainty.
\end{itemize}

For each of these benchmark networks, their true conditional probability distributions are used to sample synthetic observational data. This sampled data forms the basis for the Supervised Fine-tuning (SFT) instances and the test scenarios. As detailed in Section~\ref{subsec:augmenting_scenarios}, we applied various augmentation techniques (Name Permutation, Column Reordering, Variable Omission, and Joint Perturbations) to create a diverse and challenging corpus. This corpus is then used to construct the prompt-answer pairs for fine-tuning \CARE\ and for evaluating its performance against baseline models.

A critical aspect of our evaluation, as mentioned in Section~\ref{sec:experiments}, is that for all test scenarios, \textbf{new, unseen observational datasets} are sampled from the respective ground-truth graphs. While the graph structures themselves might be encountered during SFT (in their original or augmented forms), testing on distinct data realizations ensures that we assess the models' genuine generalization capabilities to novel data instances, rather than mere memorization of the specific training samples. This approach significantly strengthens the credibility of our performance assessments.

\section{Experiments Details}
\label{supp:experiments}

Here, we provide further details regarding the experimental setup, evaluation procedures, and implementation specifics for the results.

\subsection{Evaluation Methodology: LLM-as-Judge}
\label{suppsub:eval}
%\begin{tcolorbox} % Start of tcolorbox environment

As outlined in Section~\ref{eval_strategy}, we employ an LLM-as-a-judge approach for the automated and consistent evaluation of causal graphs predicted by different models. Specifically, \texttt{GPT-4.1-mini} is utilized as the judge. The following prompt is provided to the judge LLM for each test instance. The placeholders \texttt{gold\_answer\_str} and \texttt{model\_output\_str} are replaced with the ground-truth causal graph and the predicted graph from the model under evaluation, respectively. This structured approach ensures that the F1 score, Precision, and Recall metrics reported in the main paper are derived consistently and objectively based on the parsed edges.

\subsection{Implementation Details}
\label{suppsub:implementation}

\textbf{Baseline LLMs:} The baseline LLM evaluations (Table~\ref{tab:sec4} and Table~\ref{tab:perf_in_distribution}) for \texttt{Qwen2.5-1.5B}, \texttt{gpt-4.1-mini}, \texttt{gpt-4o-mini}, and \texttt{o4-mini} were performed using their respective publicly available APIs or Hugging Face model checkpoints at the time of experimentation (specify dates/versions if crucial). Prompts were constructed as described in Section~\ref{sec:investigation}, including observational data (for $N=50, 200$) and outputs from a suite of classical causal discovery algorithms. 

\textbf{Baseline Causal Discovery Methods:} The specific algorithms whose outputs were provided include \texttt{PC}, \texttt{GES}, \texttt{ICA-LiNGAM}, \texttt{DirectLiNGAM}, \texttt{FCI}, \texttt{GRaSP}, and \texttt{BOSS}. These were executed using standard Python libraries such as \texttt{cdt} (Causal Discovery Toolbox) and \texttt{pgmpy}. For each baseline LLM, the temperature or other generation parameters were set to ensure deterministic or near-deterministic output for reproducibility (e.g., temperature set to 0 or a very low value).

\subsection{Computational Resources}
\label{suppsub:compute}

All Supervised Fine-tuning experiments for \CARE\ and baseline LLM evaluations (where local models were run) were conducted on a compute cluster equipped with 8 NVIDIA A6000 GPUs, each with 48GB of VRAM. The specific GPU allocation per run varied depending on model size and batch configuration. For example, fine-tuning the \texttt{Qwen2.5-1.5B} model with LoRA typically utilized a single A6000 GPU. The total training time for \CARE\ across all epochs and dataset augmentations was approximately 40 hours. Evaluations involving API-based models like GPT-4 series incurred costs associated with token usage.

\subsection{Further Details on Augmentation Scenarios (Perturbation Combinations)}
\label{suppsub:aug_details}
As mentioned in Section~\ref{subsec:augmenting_scenarios}, the "Perturbation Combination" category for data augmentation, used to generate SFT instances and test scenarios, is built upon a foundational perturbation of Column Order Randomization (\textbf{A2}). This ensures the model is not sensitive to the input order of variables. The specific scenarios tested in Table~\ref{tab:perf_in_distribution} under the main "Original", "Omitted", and "Permuted" column groups implicitly include this foundational column reordering:
\begin{itemize}
    \item \textbf{Original (in Table~\ref{tab:perf_in_distribution})}: Refers to scenarios where the original semantic variable names are used, but the \emph{columns in the provided observational dataset are randomly reordered} (\textbf{A2}). The ground-truth DAG remains the same as the benchmark's original DAG.
    \item \textbf{Omitted (in Table~\ref{tab:perf_in_distribution})}: These scenarios start with Variable Omission (\textbf{A3}) to create a new, smaller set of variables and a corresponding marginalized ground-truth DAG. Then, the columns for these remaining variables in the observational dataset are \emph{randomly reordered} (\textbf{A2}).
    \item \textbf{Permuted (in Table~\ref{tab:perf_in_distribution})}: These scenarios apply Semantic Variable Name Permutation (\textbf{A1}), where the names are shuffled but the data matrix initially is not. Crucially, after this name permutation, the columns in the observational dataset (now associated with misleading names) are also \emph{randomly reordered} (\textbf{A2}). This creates a particularly challenging scenario where the model must discern true relationships despite both misleading semantic cues and arbitrary data ordering. The ground-truth DAG corresponds to the true relationships in the underlying (now reordered) data, not the permuted names.
\end{itemize}

This \textbf{consistent application of column reordering across conditions} ensures that improvements  \textbf{are not due to the model learning spurious positional clues}.

\noindent
\begin{center} 
\begin{minipage}{\textwidth}
\begin{tcolorbox}[colback=gray!5!white,colframe=gray!75!black,title=LLM-as-Judge Evaluation Prompt]
Analyze the following causal relationship information and provide a structured comparison in JSON format.

\textbf{Ground Truth Causal Relationships (Gold Standard):} gold\_answer\_str

\textbf{Model Generated Output:} model\_output\_str

\textbf{Instructions:}

\begin{enumerate}
    \item \textbf{Parse the Gold Standard:} Identify all individual directed causal edges (e.g., \texttt{"A -> B"}, \texttt{"C <- D"} which is equivalent to \texttt{"D -> C"}) from the Gold Standard section. List them using only \texttt{"->"} notation.
    \item \textbf{Parse the Model Output:} Carefully extract all causal relationships from the text block labeled "Model Generated Output". Look for patterns like \texttt{"X -> Y"}, \texttt{"X <- Y"} (which is equivalent to \texttt{"Y -> X"}), \texttt{"X causes Y"}, or \texttt{"X leads to Y"}. Consider both explicit statements like "A causes B" and structured lists (with bullet points, numbered items, etc.). Ignore introductory phrases, confidence scores, apologies, or explanations unless they define relationships. Convert all relationships to \texttt{"X -> Y"} notation - for example \texttt{"X <- Y"} should be recorded as \texttt{"Y -> X"}. Extract all relationships you find, even if they form cycles (like A->B->C->A). While proper causal graphs should be acyclic, faithfully report what the model output contains. \newline % Added newline for visual separation
    \textbf{CRITICAL INSTRUCTION:} If the text contains a section marked with \texttt{"***Answer:***"} or similar indicator (like \texttt{"Final Answer:"}, \texttt{"Answer:"}, etc.), ONLY extract edges from that section and IGNORE all other parts of the text. The relationships listed in this final answer section are the definitive edges that should be extracted, superseding any relationships mentioned earlier in the text.
    \item \textbf{Compare:} Compare the set of edges extracted from the model output against the set of edges from the Gold Standard.
    \item \textbf{Calculate Metrics:}
    \begin{itemize}
        \item \texttt{recovered\_edges}: List the edges present in \textit{both} the Gold Standard and the Model Output.
        \item \texttt{missed\_edges}: List the edges present in the Gold Standard but \textit{missing} from the Model Output.
        \item \texttt{extra\_edges}: List the edges present in the Model Output but \textit{not} in the Gold Standard.
        \item \texttt{recovered\_count}: Count of \texttt{recovered\_edges}.
        \item \texttt{missed\_count}: Count of \texttt{missed\_edges}.
        \item \texttt{extra\_count}: Count of \texttt{extra\_edges}.
    \end{itemize}
    \item \textbf{Format Output:} Return \textit{only} a single JSON object containing the following keys: \texttt{gold\_edges\_parsed}, \texttt{model\_edges\_extracted}, \texttt{recovered\_edges}, \texttt{missed\_edges}, \texttt{extra\_edges}, \texttt{recovered\_count}, \texttt{missed\_count}, \texttt{extra\_count}. Ensure edge representation is consistent (e.g., always use \texttt{"->"}).
\end{enumerate}

\textbf{Example Edge Parsing:}
\begin{itemize}
    \item \texttt{"A -> B"} is one edge.
    \item \texttt{"C <- D"} should be treated as \texttt{"D -> C"}.
    \item \texttt{"E -> F -> G"} implies edges \texttt{"E -> F"} and \texttt{"F -> G"}.
    \item \texttt{"- H -> I"} in a list implies edge \texttt{"H -> I"}.
    \item Phrases like \texttt{"X causes Y"} should be converted to \texttt{"X -> Y"}.
    \item A statement that \texttt{"X influences Y and Z"} implies both \texttt{"X -> Y"} and \texttt{"X -> Z"}.
\end{itemize}
\end{tcolorbox} % End of tcolorbox environment
%\end{tcolorbox} % End of tcolorbox environment
\end{minipage}
\end{center} % End of optional centering

\section{Limitations and Discussions}
\label{sec:limitations}

While \CARE\ demonstrates strong performance with a 1.5B parameter model, we identify several directions for future exploration and potential improvements:
\begin{itemize}
    \item \textbf{Impact of Base Model Scale and Computational Resources:} Our current study utilized the Qwen2.5-1.5B model, guided by available computational resources. Applying the \CARE\ SFT methodology to larger base LLMs (e.g., 7B, 70B, or beyond) is anticipated to yield further substantial gains in causal reasoning capabilities, leveraging their inherently greater knowledge and capacity. Exploring this scaling is a promising direction for future work.
    \item \textbf{Scalability to Extremely Large and Complex Graphs:} While effective on benchmarks like ALARM (37 nodes), investigating \CARE's performance on problems with extremely large numbers of variables (hundreds or thousands), especially when combined with larger base models as mentioned above, remains an important area. This includes managing the complexity of algorithmic outputs and potential LLM context window considerations.
    \item \textbf{Resource Considerations for SFT and Data Curation:} Effective SFT, as employed by \CARE, benefits from a diverse corpus of training instances. The process of generating and curating these augmented datasets, alongside the general computational costs of SFT (even with PEFT), are practical considerations for adapting \CARE\ to new, specialized domains.
\end{itemize}

\end{document}